\documentclass{article}

\PassOptionsToPackage{square, numbers}{natbib}
\usepackage[preprint]{neurips_2024}
\usepackage[utf8]{inputenc} % allow utf-8 input
\usepackage[T1]{fontenc}    % use 8-bit T1 fonts
\usepackage{hyperref}       % hyperlinks
\usepackage{url}            % simple URL typesetting
\usepackage{booktabs}       % professional-quality tables
\usepackage{amsfonts}       % blackboard math symbols
\usepackage{amsthm}
\usepackage{amsmath}
\usepackage{amssymb}
\usepackage{nicefrac}       % compact symbols for 1/2, etc.
\usepackage{microtype}      % microtypography
\usepackage{xcolor}         % colors

\usepackage{multirow}
\usepackage{makecell}
\usepackage{graphicx}
\usepackage{subcaption}
\usepackage{wrapfig}
\usepackage{appendix}
\usepackage{arydshln}

\usepackage{pifont}

\usepackage{algorithm}
\usepackage{algorithmic}
\usepackage{listings}
\newcommand{\tablestyle}[2]{\setlength{\tabcolsep}{#1}\renewcommand{\arraystretch}{#2}\centering\footnotesize}
\title{AVESFormer: Efficient Transformer Design for Real-Time Audio-Visual Segmentation}

\author{%
  Zili Wang$^{1,2}$\quad 
  Qi Yang$^{1,2}$\quad 
  Linsu Shi$^3$\quad
  Jiazhong Yu$^3$\quad\\
  \textbf{Qinghua Liang}$^3$\quad
  \textbf{Fei Li}$^3$\quad
  \textbf{Shiming Xiang}$^{1,2}$\\ \\
  $^1$School of Artificial Intelligence, University of Chinese Academy of Sciences (UCAS)\\
  $^2$Institute of Automation, Chinese Academy of Sciences (CASIA)\\
  $^3$China Tower Corporation Limited
}

\begin{document}

\maketitle

\begin{abstract}
  Recently, transformer-based models have demonstrated remarkable performance on audio-visual segmentation (AVS) tasks.
  However, their expensive computational cost makes real-time inference impractical.
  By characterizing attention maps of the network, we identify two key obstacles in AVS models: 1) attention dissipation, corresponding to the over-concentrated attention weights by Softmax within restricted frames, and 2) inefficient, burdensome transformer decoder, caused by narrow focus patterns in early stages.
  In this paper, we introduce \textbf{AVESFormer}, the first real-time \textbf{A}udio-\textbf{V}isual \textbf{E}fficient \textbf{S}egmentation transformer that achieves fast, efficient and light-weight simultaneously.
  Our model leverages an efficient prompt query generator to correct the behaviour of cross-attention. 
  Additionally, we propose ELF decoder to bring greater efficiency by facilitating convolutions suitable for local features to reduce computational burdens.
  Extensive experiments demonstrate that our AVESFormer significantly enhances model performance, achieving 79.9\% on S4, 57.9\% on MS3 and 31.2\% on AVSS, outperforming previous state-of-the-art and achieving an excellent trade-off between performance and speed.
  Code can be found \href{https://github.com/MarkXCloud/AVESFormer.git}{here}.
\end{abstract}

\begin{figure}[h]
    \centering
    \includegraphics[width=1.0\linewidth]{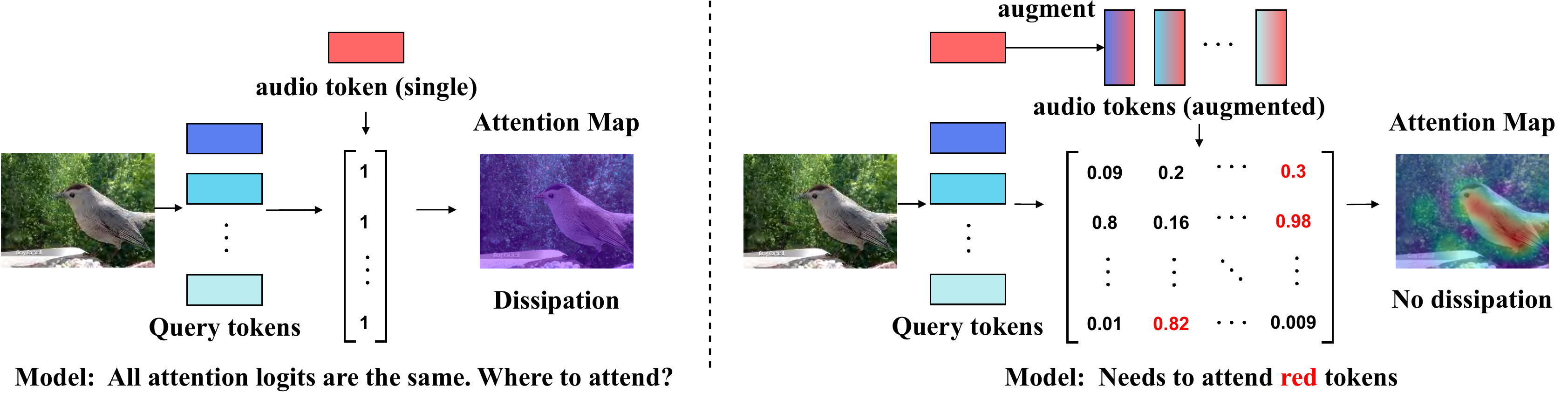}
    \caption{Illustration of attention dissipation. The cross-attention matrix fails to distinguish different tokens (left). One potential solution is to expand the audio feature into several tokens (right).}
    \label{fig:dissipation}
\end{figure}
\vspace{-1.35em}

\section{Introduction}
Audio-Visual Segmentation (AVS)~\cite{zhou2022audio} have emerged as a novel multi-modality task that plays a crucial role in robot sensing, video surveillance and other scenarios. It aims to segment fine-grained pixel-level sounding objects with corresponding audio-visual modalities. However, existing AVS methods~\cite{gao2024avsegformer,li2023catr,mao2023ecmvae,liu2023autr,mao2023diffusionavs,huang2023aqformer,liu2023avsc,hao2024avsbg,liu2023bavs,li2023towards} primarily focus on improving performance, often at a high cost of models size and computational overhead. Such heavy computational cost renders them unsuitable for applications with real-time requirements.

Recently, transformer-based AVS models have brought significant performance improvements with cross-attention and its variants serving as audio-visual fusion module~\cite{gao2024avsegformer,yang2023cooperation,huang2023aqformer,li2023towards,liu2023autr,liu2023bavs,li2023catr}. Beginning with AVSBench~\cite{zhou2022audio}, temporal pixel-wise audio-visual interaction (TPAVI)~\cite{zhou2022audio,zhou2023avss} is proposed to inject audio guidance from all video clips. However, such method is unnatural since the sound source may change during the clip.
AVSegFormer~\cite{gao2024avsegformer} employs channel attention mixer (CHA) to guide visual channels with audio features. Nevertheless, channel attention may be dominated by visual features and surpass the audio representation~\cite{zhou2022acloserlook}.
Contrastive Audio-Visual Pairing (CAVP)~\cite{zhou2022acloserlook} approximates Softmax function with Sigmoid, suggesting it could highlight critical regions. Nonetheless, approximated attention does not hold the same power as attention~\cite{li2022efficientformer}. 

\begin{wrapfigure}[13]{rh}{6.13cm}
    \centering
    \includegraphics[width=1.0\linewidth]{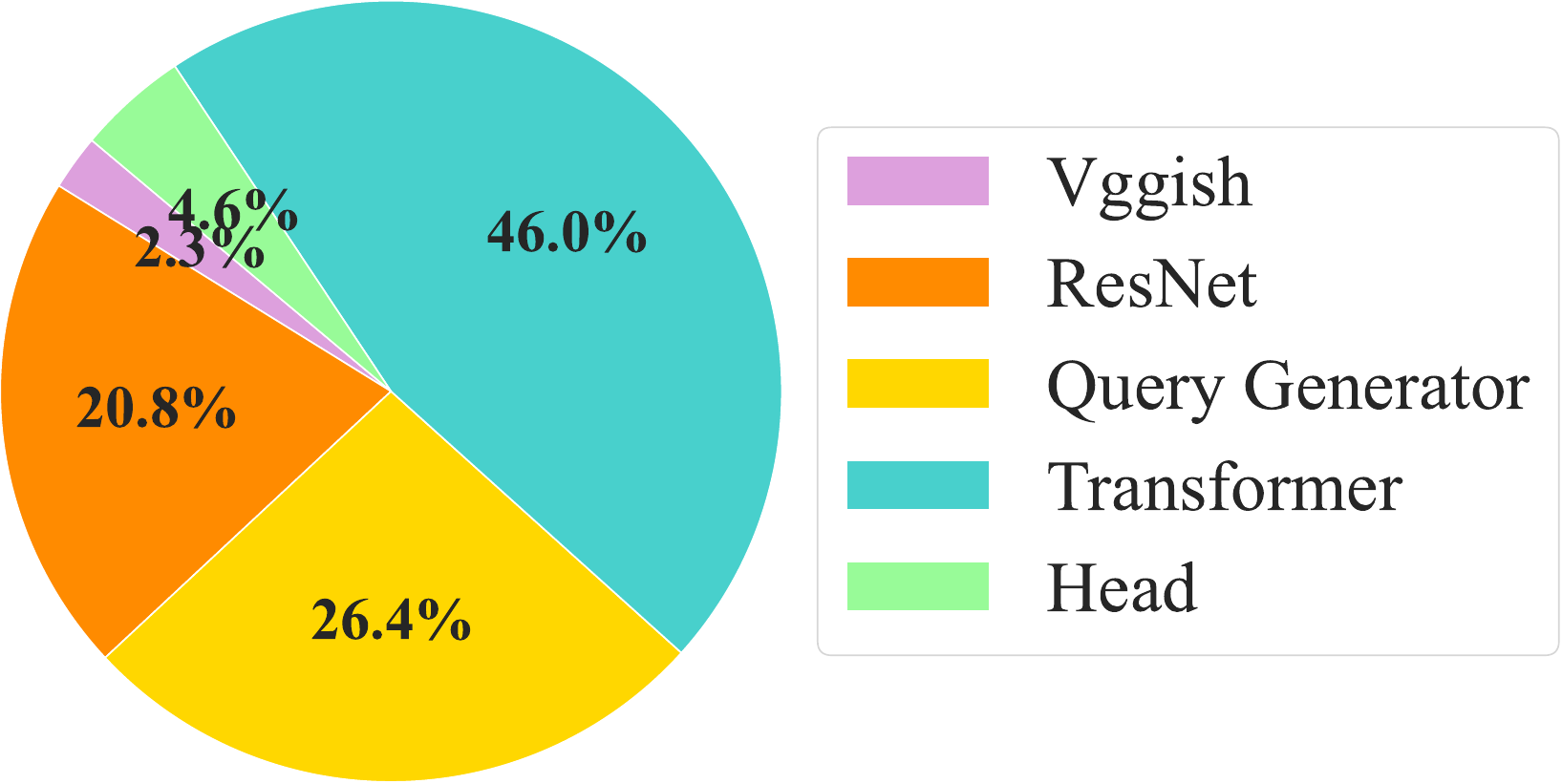}
    \caption{Runtime profiling of the AVSegFormer~\cite{gao2024avsegformer}. Inference time is dominated by transformer architecture.}
    \label{fig:profile}
\end{wrapfigure}

Despite their strong performance, the application of AVS model on real-time field is still difficult because the computational overheads and model efficiency are often neglected. Our observation identifies two primary issues that prevent the AVS model from the real-time area: (1) \textbf{Attention Dissipation}, an issue not explored in previous studies, where cross-attention matrix vanishes during modality fusion process in existing methods~\cite{gao2024avsegformer,li2023catr,liu2023autr}, hindering them from distinguishing audio-visual corresponding region, as shown in Figure~\ref{fig:dissipation}. (2) \textbf{Inefficient Decoder} tends to capture narrow local features at early stages with cross-attention, resulting in short-range pattern utilization, which is not a desired behaviour of attention. These inefficient operations not only fail to build long-range dependencies, but also constitute the bottleneck of inference runtime. As shown in Figure~\ref{fig:profile}, the runtime proportion of the transformer, including the query generator, can exceed 70\% of the total.

To this end, we settle into a better explanation of attention dissipation and seek to reduce transformer decoder overhead while enhancing its efficiency. We analyze that attention dissipation is derived from over-concentrated distribution across multiple elements of the attention weights after Softmax within restricted frames. To address this issue, \textbf{P}ompt \textbf{Q}uery \textbf{G}enerator (\textbf{PQG}) is adopted to process the audio feature in a prompt manner. This approach rebuilds the distinguishing capability of cross-attention, effectively eliminating attention dissipation. For improving decoder efficiency, a novel \textbf{E}ar\textbf{L}y \textbf{F}ocus (\textbf{ELF}) decoder is introduced. Specifically, convolution blocks are conducted in the early transformer decoder stages. This modification is more suitable to capture local features in contrast to attention while reducing the computational cost of the latter. Our method proves to be faster and more efficient than relying solely on cross-attention throughout the entire decoder.

\begin{figure}[t]
    \centering
    \includegraphics[width=1.0\textwidth]{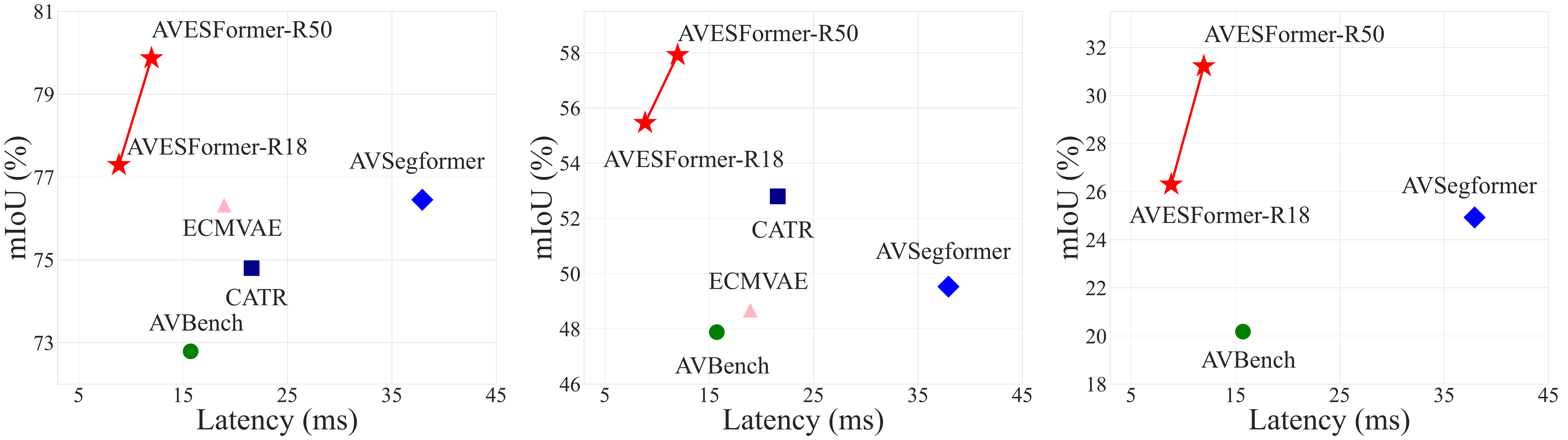}
    \caption{\textbf{mIoU (\%) vs. Inference Latency (ms)} on S4 (left), MS3 (middle) and AVSS (right) compared with other popular methods. Latency is measured on a single Nvidia RTX 3090 GPU. AVESFormer achieves the best trade-off between performance and inference speed.}
    \label{fig:intro}
\end{figure}

In this work, we introduce \textbf{AVESFormer}, an \textbf{A}udio-\textbf{V}isual \textbf{E}fficient \textbf{S}egmentation transformer, which achieves fast, efficient and light-weight simultaneously. To the best of our knowledge, AVESFormer is the first real-time transformer model for AVS tasks. AVESFormer addresses the critical issue of attention dissipation through prompt query generator and reduces inference runtime with efficient ELF decoder. As shown in Figure~\ref{fig:intro}, comprehensive experiments demonstrate that AVESFormer achieves state-of-the-art performance-latency trade-off, outperforming AVSegFormer~\cite{gao2024avsegformer} (+3.4\% on S4, +8.4\% on MS3, +6.3\% on AVSS) while using $20\%$ less parameters and being $3\times$ faster.

Our contributions can be summarized as follows:
\begin{itemize}
\item We discover the attention dissipation phenomenon in the cross-attention fusion process. To address this, we propose a novel prompt audio query generator that corrects its behaviour and establishes a reliable representation capability of audio-visual fusion.

\item We identify insufficient audio-visual fusion in the early stages of the transformer decoder. Thus we adopt ELF decoder, which reduces computational cost and promotes efficient audio-visual fusion in deeper stages.

\item Our method achieves state-of-the-art w.r.t. the trade-off between performance and inference speed on challenging AVSBench-Object and AVSBench-Semantic datasets.
\end{itemize} 
%%%%%%%%%%%%%%%%%%%%%%%%%%%%%%%%%%%%%%%%%%%%%%%%%%%%%%%%%%%%
\section{Related Work}

\subsection{Audio-Visual Segmentation}
Audio-Visual Segmentation (AVS) is a more fine-grained and complicated task than sound source localization (SSL)~\cite{chen2021localizing,hu2020discriminative,qian2020multiple} as it aims to locate the sounding object and show pixel-level predictions. In recent years, AVS has attracted significant attention from researchers.
AVSBench~\cite{zhou2022audio} is the first to propose audio-visual segmentation benchmark, introducing temporal pixel-wise audio-visual interaction (TPAVI) module to facilitate interaction between audio-visual information.
AVSegFormer~\cite{gao2024avsegformer} is the first to develop a novel transformer architecture for AVS. They introduce audio queries into the transformer decoder to attend to corresponding visual features.
CATR~\cite{li2023catr} performs bidirectional combinatorial dependence fusion to fully enhance spatial-temporal dependencies.
CAVP~\cite{zhou2022acloserlook} incorporates contrastive loss into audio-visual semantic segmentation with positive and negative pairs and uses larger resolution with extra data to reach higher performance.
Unlike these methods, this paper focuses on real-time end-to-end inference scenario of AVS model and provides a detailed analysis of attention dissipation and decoder efficiency.

\subsection{Transformer in Semantic Segmentation}
In recent years, transformer architecture has significantly influenced semantic segmentation.
DPT~\cite{ranftl2021dpt} designs a transformer-based encoder-decoder architecture for dense prediction tasks.
SETR~\cite{zheng2021setr} shows impressive results by modelling segmentation as a sequence-to-sequence task.
SegFormer~\cite{xie2021segformer} introduces a hierarchical transformer encoder and an all-MLP decoder to improve the network efficiency.
MaskFormer~\cite{cheng2021maskformer} and Mask2Former~\cite{cheng2022mask2former} modify segmentation in 
a set prediction paradigm, predicting a set of binary masks and assigning a single category to each one.
However, these models are unsuitable for real-time segmentation tasks due to their heavy computational burden.
RTFormer~\cite{wang2022rtformer} introduces GPU-Friendly attention and arranges low- and high-resolution branches in a stepped layout to make full use of global context.
SeaFormer~\cite{wan2023seaformer} employs squeeze axial attention to reduce the computation burden of self-attention while maintaining the local details.
These methods have significantly advanced semantic segmentation. Considering the tight bond between segmentation and AVS tasks, these approaches have provided substantial inspiration for our work.

\subsection{Efficient Vision Transformer}
ViT~\cite{dosovitskiy2020vit} and its variants~\cite{liu2021swin,touvron2021deit,wang2022pvt} have demonstrated significant improvements in computer vision. However the high computational cost makes them inferior to CNN in real-time inference scenario. To mitigate this gap, previous works attempt to design more efficient architectures to reduce computational burden. MobileViT~\cite{mehta2021mobilevit} combines CNN and ViT by integrating global feature fusion of transformer in CNN. MobileFormer~\cite{chen2022mobileformer} bridges MobileNet~\cite{howard2017mobilenet} and ViT in a parallel design to leverage advantages from both architectures. EfficientFormer~\cite{li2022efficientformer} finds insufficient operations in transformer and slims the model size in a latency-driven manner. LVT~\cite{xiao2021lvt} adopts dilated convolution in attention mechanisms to enhance model performance and efficiency. LIT~\cite{pan2022lit} gives a more detailed analysis of self-attention heads and applies MLP to build local dependencies. EfficientViT~\cite{cai2022efficientvit_mit} proposed to aggregate multi-scale features via small-kernel convolutions. These methods have made contributions to the development of fast and efficient ViT architectures. We benefit greatly from their contributions to the analysis of AVS tasks.

\begin{figure}[t]
    \centering
    \includegraphics[width=1.0\textwidth]{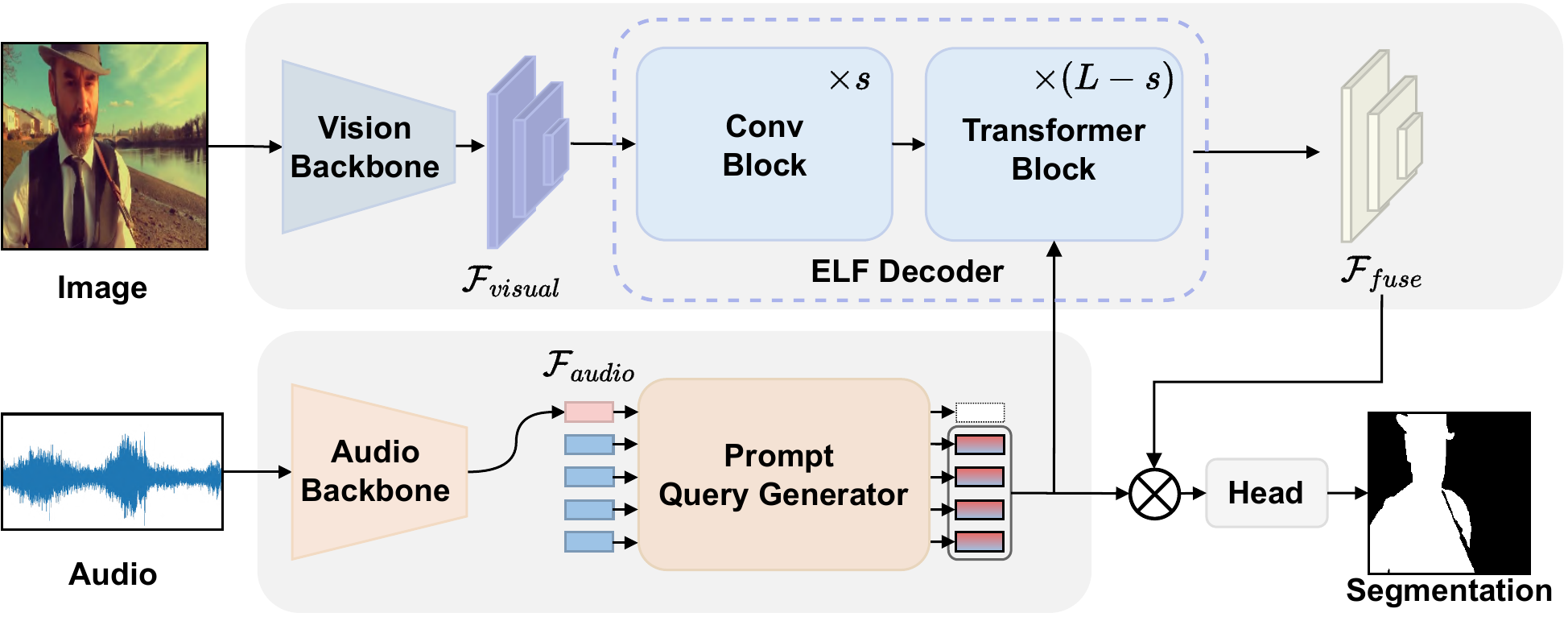}
    \caption{The overview of AVESFormer. Audio and visual backbones extract corresponding features. The prompt query generator addresses the attention dissipation problem by inserting the audio feature on top of a set of learnable parameters to generate audio-conditioned queries. The ELF decoder processes local features using convolution blocks in the early stages. Finally, the transformer blocks interact with high-level audio-visual features to generate fused features.}
    \label{fig:architecture}
\end{figure}

\section{Method}
In this section, we first describe the theoretical analysis of the attention dissipation phenomenon. Then, we elaborate the detailed architecture and components of the proposed AVESFormer.

\subsection{Attention Dissipation}\label{sec:dissipation}
In real-time AVS scenario, visual feature $\mathcal{F}_{visual}\in\mathbb{R}^{c\times h\times w}$ and audio feature $\mathcal{F}_{audio}\in\mathbb{R}^{1\times c}$ are given at the same moment. The former is usually split into patches $\mathcal{P}_{visual}\in\mathbb{R}^{N\times c}$ where $N=h\times w$ in the cross attention mechanism. Prevailing methods~\cite{gao2024avsegformer,zhou2022acloserlook,li2023catr,li2023towards} usually fuse the two features to build reliable correspondence between audio-visual modalities, as shown on the left panel of Figure~\ref{fig:dissipation}. Let us denote $q_i, k, v\in\mathbb{R}^{1\times c}$ as row vectors for $i\in[1,2,\dots,N]$, with $\mathcal{P}_{visual}=[q_i]_{N\times c}$ and $\mathcal{F}_{audio}=k=v$. The cross-attention fusion can be represented as follows:
\begin{align}
    \mathcal{O}&=\text{Softmax}(\mathcal{P}_{visual}\mathcal{F}_{audio}^T)\mathcal{F}_{audio},\\[5pt]
    o_i&=\sum_ja_{i,j}v_j=\frac{\sum_je^{q_ik_j^T}v_j}{\sum_je^{q_ik_j^T}},\label{eq:smallo}
\end{align}
where $\mathcal{O}=\begin{bmatrix}
        o_i
    \end{bmatrix}\in\mathbb{R}^{N\times c}$ and $j$ stands for the row index of $\mathcal{F}_{audio}$.
The scale factor $\sqrt{d}$ in Softmax as well as linear transformation matrices of $W^Q$, $W^K$ and $W^V$~\cite{vaswani2017attention} are omitted for the sake of simplicity without affecting the conclusion.

However, $\mathcal{F}_{audio}$ is an 1-dimensional vector within a single frame, which makes $k_j=k$ and $v_j=v$. Based on this hypothesis, we substitute $j=1$ into Equation (\ref{eq:smallo}) to obtain:
\begin{align}
    o_i&=\frac{e^{q_ik^T}v}{e^{q_ik^T}}
    =v.
\end{align}
The final output of cross-attention fusion can be written as:
\begin{align}
    \mathcal{O}&=\text{Softmax}(\begin{matrix}
    \{q_ik^T\}_{ij}\\
    \end{matrix})v=\mathbf{1}_{N\times 1}\mathcal{F}_{audio}=\begin{bmatrix}
        \mathcal{F}_{audio}\\
    \end{bmatrix}_{N\times c}.\label{eq:dissipation}
\end{align}
From Equation~(\ref{eq:dissipation}), the cross-attention fusion turns into a simple replication of the audio feature, as illustrated on the right panel of Figure~\ref{fig:dissipation}. In this process, the attention weights turn out to be over-concentrated after Softmax over 1-dimensional keys. The phenomenon revealed in Equation~(\ref{eq:dissipation}), termed \textbf{Attention Dissipation}, significantly harms the capability of distributing attention on multi-modality representation,  thus constraining the effectiveness of the attention mechanism~\cite{gao2024avsegformer,zhou2022acloserlook}. Modifications to the audio features are necessary to correct the behaviour of cross-attention. One potential solution is to expand the amount of audio tokens. See Appendix~\ref{app:dissipation_detail} for more proof details.
\subsection{AVESFormer Architecture}
We now introduce the overall architecture of AVESFormer, including audio-visual backbones, prompt query generator, early focus decoder and loss function, as shown in Figure~\ref{fig:architecture}.
\paragraph{Visual Backbone.} Initially, audio-visual features are extracted by corresponding backbones.
For a single frame $x_{visual}\in\mathbb{R}^{3\times H\times W}$, where $H$ and $W$ stand for the
height and width of the image, hierarchical visual features are extracted as follows:
\begin{align}
  \mathcal{F}_{visual}&=\{\mathcal{F}_1,\mathcal{F}_2,\mathcal{F}_3,\mathcal{F}_4\},
\end{align}
where $\mathcal{F}_i\in\mathbb{R}^{c_i\times\frac{H}{2^{i+1}}\times\frac{H}{2^{i+1}}}$, $i\in\{1,2,3,4\}$ represents features at different scale with channel $c_i$. 

\paragraph{Audio Backbone.} Meanwhile, the audio signal in the video with time duration $T$ is resampled to yield a 16kHz mono output $A_{mono}\in\mathbb{R}^{N_{samples}\times 96 \times 64}$, where $N_{samples}$ stands for the amount of sampling points. Then, $A_{mono}$ is converted into Mel-spectrum 
$A_{mel}\in\mathbb{R}^{T\times 96\times 64}$ by short-time 
Fourier transform. Finally we put $A_{mel}$ into pre-trained audio backbone Vggish~\cite{hershey2017vggish} to extract $T$ audio features, each one is denoted as $\mathcal{F}_{audio}\in\mathbb{R}^{1\times D}$, where $D$ is the audio feature dimension.

\begin{wrapfigure}[20]{rh}{5.5cm}
    \centering
    \includegraphics[width=0.96\linewidth]{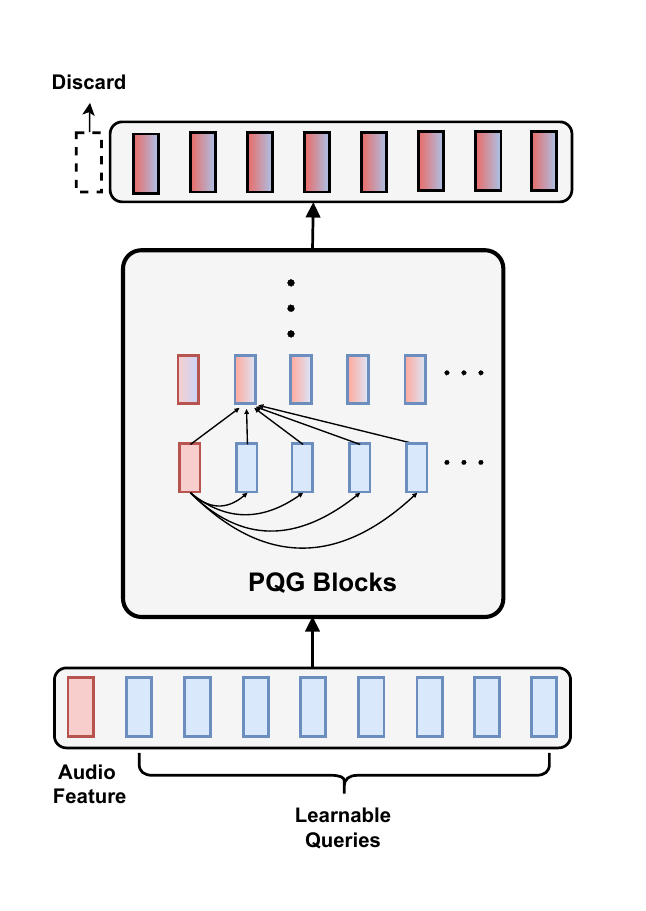}
    \caption{Illustration of prompt query generator. The audio feature is treated as prompt and discarded in output.}
    \label{fig:query_generator}
\end{wrapfigure}
\paragraph{Prompt Query Generator.} To mitigate the attention dissipation discussed in Section~\ref{sec:dissipation}, a novel prompt query generator (PQG) is proposed to expand audio query rather than replicating, as depicted in Figure~\ref{fig:query_generator}. The audio feature at a single frame is regarded as a \textbf{prompt}~\cite{liu2023prompt} and inserted into a set of learnable queries $Q_{learn}\in\mathbb{R}^{N_{q}\times D}$:
\begin{align}
    Q^\dagger=[\mathcal{F}_{audio}|Q_{learn}]\in\mathbb{R}^{(N_{q}+1)\times D},
\end{align}
where $[ \cdot | \cdot ]$ denotes concatenation and $N_q$ denotes query amount. Then, self-attention is performed between audio features and learnable queries. In the self-attention process, 
the attention matrix $QK^T$ can be written as follows:
\renewcommand{\arraystretch}{1.75}
\begin{align}
    QK^T&=Q^\dagger Q^{\dagger T}\\[5pt]
    &=[\mathcal{F}_{audio}|Q_{learn}][\mathcal{F}_{audio}|Q_{learn}]^T\\[5pt]
    &=
    \left[\begin{array}{c:c}
         \mathcal{F}_{audio}\mathcal{F}_{audio}^T&\mathcal{F}_{audio}Q_{learn}^T\\
        \hdashline
        Q_{learn}\mathcal{F}_{audio}^T&Q_{learn}Q_{learn}^T
    \end{array}\right].
\end{align}
\renewcommand{\arraystretch}{1}

Afterwards, $Q^\dagger$ generates augmented audio features with relativity from the original feature. Lastly, the original audio token at the output end is discarded to obtain $\mathcal{F}_{gen}\in\mathbb{R}^{N_{q}\times D}$. Note that PQG enhances the diversity of audio features and corrects the behaviour of cross-attention.

\begin{figure}[t]
    \centering
    \includegraphics[width=0.75\linewidth]{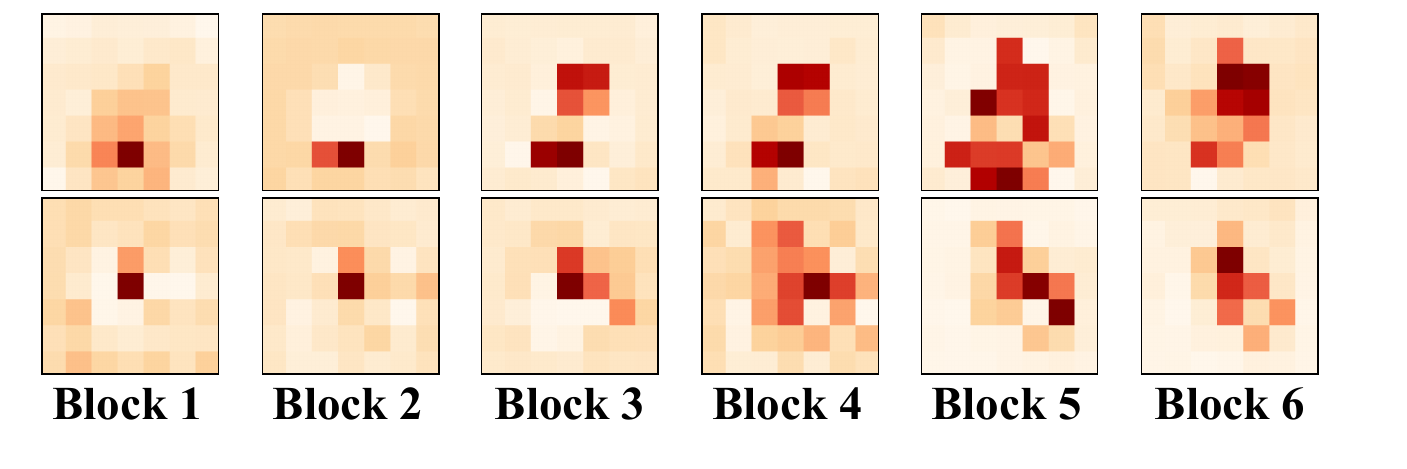}
    \caption{Attention probabilities of different blocks in fully transformer decoder. Each map shows the attention probability of the audio query to all visual patches. Maps are averaged along all heads. Each row indicates a test sample. Dark red indicates higher attention probability and shallow orange indicates lower attention probability. See Appendix~\ref{app:qualitative_early.} for more details.}
    \label{fig:visual_early}
\end{figure}

\paragraph{Early Focus Decoder.} Despite its powerful representation ability, the transformer decoder remains the main bottleneck of runtime, as shown in Figure~\ref{fig:profile}. Previous works on ViTs~\cite{xiao2021lvt,pan2022lit} suggest that early stages of self-attention tend to be inefficient because they predominantly focus on local patterns, leading to wasted long-range modelling capability.  In contrast, deeper stages mainly capture long-range, high-level semantics. In this work, we visualize the audio-visual cross-attention patterns, as shown in Figure~\ref{fig:visual_early}. In the early stages, audio features generate narrow local 
 visual responses on attention maps. As the stage goes deeper, the attention region enlarges gradually. In the last two stages, it forms shaped and fine-grained regions suitable for segmentation. Therefore, we propose a novel early focus (ELF) decoder. Since the early stage primarily captures local patterns, attention to high computational cost is replaced by convolution to capture local semantics. In early decoder stage $l$, visual feature $\mathcal{F}_{visual}$ is processed by convolution:
\begin{align}
    \mathcal{F}_{visual}^{l+1}=\text{LN}(\mathcal{F}_{visual}^l+\text{Conv}(\mathcal{F}_{visual}^l)),
\end{align}
where LN denotes LayerNorm~\cite{ba2016layernorm} and Conv is composed of RepBlock~\cite{ding2021repvgg}. In deeper stages, we split $\mathcal{F}_{visual}$ into visual patches $\mathcal{P}_{visual}$~\cite{dosovitskiy2020vit} to perform cross-attention with $\mathcal{F}_{gen}$ from PQG:
\begin{align}
    \mathcal{P}_{visual}^{l+1}&=\text{LN}(\mathcal{P}_{visual}^l+\text{CA}(\mathcal{P}_{visual}^l,\mathcal{F}_{gen},\mathcal{F}_{gen})),
\end{align}
where CA denotes multi-head cross-attention and $\text{CA}(Q,K,V)=\text{Softmax}(QK^T)V$. The ELF decoder eliminates the computational burden brought by wasted attention operations but still maintains the original module function to extract local features. By incorporating our ELF decoder, we find more performance-computation efficiency enhancement in our model.

\paragraph{Loss Function.}Following MaskFormer~\cite{cheng2021maskformer}, we employ IoU loss and Dice~\cite{milletari2016dice} loss to provide supervision between the predicted mask $\hat{\mathcal{M}}$ and ground truth $\mathcal{M}$. 
The IoU loss $\mathcal{L}_{\text{IoU}}$ measures the intersection over union between prediction and ground truth.
Moreover, Dice loss $\mathcal{L}_{\text{Dice}}$ is set to obtain additional supervision information.
Since the foreground proportion in the AVS task is relatively small within the entire image, Dice loss
could force the model to focus on the target region and suppress the impact of background interference. 
Besides, we employ auxiliary loss $\mathcal{L}_{\text{aux}}$ for more fine-grained segmentation. The intermediate feature from convolution blocks of ELF decoder $\mathcal{F}_{ELF}\in\mathbb{R}^{c\times h \times w}$ is introduced to calculate $\mathcal{L}_{\text{aux}}$. Suppose $\mathcal{M}_f$ represents the foreground mask, $\mathcal{L}_{\text{aux}}$ can be written as:
\begin{align}
    \mathcal{L}_{\text{aux}}(\mathcal{F}_{ELF}, \mathcal{M}_f)&=\frac{1}{c}\sum_{i=1}^c\mathcal{L}_{\text{Dice}}(\mathcal{F}_{ELF}^i,\mathcal{M}_f).
\end{align}
The total segmentation loss can be written as:
\begin{align}
    \mathcal{L}&=\lambda_{\text{IoU}}\mathcal{L}_{\text{IoU}}(\hat{\mathcal{M}},\mathcal{M})+\lambda_{\text{Dice}}\mathcal{L}_{\text{Dice}}(\hat{\mathcal{M}},\mathcal{M})+\lambda_{\text{aux}}\mathcal{L}_{\text{aux}}(\mathcal{F}_{ELF}, \mathcal{M}_f),
\end{align}
where $\lambda_{\text{IoU}}$, $\lambda_{\text{Dice}}$ and $\lambda_{\text{aux}}$ are hyperparameters. See Appendix~\ref{app:exp_detail} for more details.

\section{Experiments}
\paragraph{Dataset.} 
We evaluate our method on the AVSBench dataset~\cite{zhou2022audio,zhou2023avss}, which is composed of AVSBench-Object and AVSBench-Semantic. 
AVSBench-Object is designed for audio-visual segmentation tasks with pixel-level annotations. Videos are sourced from YouTube, cropped into 5 seconds, and sampled at one frame per second to compose the image data. There are two subsets in AVSBench-Object: single sound source segmentation (S4) subset and multiple sound source segmentation (MS3) subset. The S4 subset contains 4,932 videos: 3,452 for training, 740 for validation and 740 for testing. The labels contain 23 categories, including humans, vehicles, animals and kinds of instruments. Note that annotations in S4 training set is only given in the first frame. Meanwhile, MS3 subset is composed of multiple sound sources, including 424 videos, 286 for training, 64 for validation and 64 for testing. MS3 shares the same categories as S4. 
AVSBench-Semantic is an expanded version of AVSBench-Object, providing additional semantic masks to facilitate audio-visual semantic segmentation (AVSS). Videos in AVSBench-Semantic extend up to 10 seconds with 10 frames per video to compose the image data. Moreover, 70 categories are annotated in 11,356 videos: 8,498 for training, 1,304 for validation and 1,554 for testing.
\paragraph{Implementation Details.}\label{sec:implementation} We conduct our experiments with PyTorch. Our model is trained on NVIDIA RTX 3090 GPU. We employ AdamW~\cite{loshchilov2017adamw} as optimizer with batch size 16 and learning rate of 0.0005 for S4 as well as MS3 while batch size 8 and learning rate of 0.0001 for AVSS. All images are resized into $224\times 224$.
From the aspect of real-time inference, we employ ResNet-50 and ResNet-18~\cite{he2016resnet} pre-trained on ImageNet~\cite{russakovsky2015imagenet} as our visual backbones. Considering Pyramid Vision Transformer (PVT-v2)~\cite{wang2022pvt} is unsuitable for real-time applications, we do not adopt it as the visual backbone. We employ Vggish~\cite{hershey2017vggish} pre-trained on AudioSet~\cite{gemmeke2017audioset} to
encode audio input. The audio backbone is frozen during the training. The embedding dimensions of both encoders are set to 256. Transformer decoder comes up with multi-scale deformable attention (MSDeform)~\cite{zhu2020deformabledetr} followed by self-attention~\cite{vaswani2017attention} and FFN. See Appendix~\ref{app:exp_detail} for more experimental details.

\paragraph{Evaluation Metrics.} Following~\cite{zhou2022audio}, we adopt Jaccard index $\mathcal{J}$ and F-score $\mathcal{F}$ to evaluate. $\mathcal{J}$ indicates the mean intersection over union (mIoU)~\cite{everingham2015miou} between segmentation prediction and ground truth. $\mathcal{F}$ measures the precision and recall by $\mathcal{F}=\frac{(1+\beta^2\times \text{precision}\times\text{recall})}{\beta^2\times \text{precision}+\text{recall}}$, where $\beta^2=0.3$.

\begin{table}[t]
\begin{center}
  \caption{Comparison with state-of-the-art methods on the AVS benchmark.
  All methods are evaluated on three AVS sub-tasks (S4, MS3 and AVSS).
  The evaluation metrics are mIoU and F-score. \#Params refers to the number of parameters. FPS is reported on a single NVIDIA RTX 3090 GPU.
  * means the parameters of audio backbone Vggish~\cite{hershey2017vggish} are included. 
  }
  \resizebox{\textwidth}{!}{
  \begin{tabular}{ll cc c cc c cc c c}
    \toprule
    \multirow{2}{*}{Method} & \multirow{2}{*}{Backbone} & \multicolumn{2}{c}{S4} & & \multicolumn{2}{c}{MS3} & & \multicolumn{2}{c}{AVSS}  & {\#Params$^*$} & \multirow{2}{*}{FPS}\\
    \cmidrule{3-4} \cmidrule{6-7} \cmidrule{9-10} 
     & & $\mathcal{J}$ & $\mathcal{F}$ & & $\mathcal{J}$ & $\mathcal{F}$&  & $\mathcal{J}$ &  $\mathcal{F}$ & (M)& \\
     \midrule
     LVS~\cite{chen2021lvs}& ResNet-18 & 38.0 & 51.0 & & 29.5 & 33.0 & & -& - & -& -\\
     MSSL~\cite{qian2020mssl}& ResNet-18 & 44.9 & 66.3 & & 26.1 & 36.3 & & -& - & -& -\\
     3DC~\cite{mahadevan20203dc}& ResNet-152 & 57.1 & 75.9 & & 36.9 & 50.3 & & 17.3 & 21.6 & -& -\\
     SST~\cite{duke2021sst}& ResNet-101 & 66.3 & 80.1 & & 42.6 & 57.2 & & -& - & -& -\\
     AOT~\cite{yang2021aot}& Swin-B & - & - & & - & - & & 25.4 & 31.0 & -& -\\
     iGAN~\cite{mao2021igan}& Swin-T & 61.6 & 77.8 & & 42.9 & 54.4 & & -& - & -& -\\
     LGVT~\cite{zhang2021lgvt}& Swin-T & 74.9 & 87.3 & & 40.7 & 59.3 & & -& - & -& -\\
    \midrule
    AVSBench~\cite{zhou2022audio}&\multirow{10}{*}{ResNet-50}&72.8 &84.8 &  & 47.9 & 57.8&  & 20.2 &25.2 &163 & 63.6 \\

    CATR~\cite{li2023catr}& & 74.8  &86.6&   & 52.8 & 65.3&  & - & - & 177 &  46.4 \\

    DiffusionAVS~\cite{mao2023diffusionavs}& & 75.8  &86.9 &  & 49.8 & 62.1&  & - & - &-& - \\

    ECMVAE~\cite{mao2023ecmvae}& & 76.3  &86.5&   & 48.7 & 60.7&  & - & - & 162 & 52.8 \\

    AuTR~\cite{liu2023autr}& & 75.0  &85.2&   & 49.4 & 61.2&  & - & - &-& - \\

    AQFormer~\cite{huang2023aqformer}& & 77.0  & 86.4 &  & 55.7 & 66.9&  & - & - &-& - \\

    AVSC~\cite{liu2023avsc}& & 77.0  & 85.2&  & 49.6 & 61.5&  & - & - &-& - \\

    AVSegFormer~\cite{gao2024avsegformer}& &76.5  &85.9&   & 49.5 & 62.8&  & 24.9 & 29.3 &151 & 26.4 \\

    AVSBG~\cite{hao2024avsbg}& & 74.1  & 85.4&  & 45.0 & 56.8 &  & - & - & - & - \\

    BAVS~\cite{liu2023bavs}& &78.0  &85.3 &  & 50.2 & 62.4&  & 24.7 & 29.6 &118 & - \\
    \midrule
    \multirow{2}{*}{AVESFormer (ours)}&ResNet-18& 77.3 & 87.5 &  & 55.5 & 65.1 &  & 26.3 & 31.8 &\textbf{108}&\textbf{113.0}  \\
     &ResNet-50& \textbf{79.9}  & \textbf{89.1} &   & \textbf{57.9} & \textbf{68.7}&  & \textbf{31.2} & \textbf{36.8} & 127 & 83.5  \\
  \bottomrule
\end{tabular}
}
  \label{tab:main_result}
  \end{center}
\end{table}

\subsection{Comparison with State-of-the-arts}
Comprehensive experiments have been conducted on AVSBench-Object and AVSBench-Semantic datasets alongside other methods. As shown in Table~\ref{tab:main_result}, our AVESFormer exhibits the state-of-the-art performance-speed trade-off among all models. Specifically, AVESFormer surpasses previous methods w.r.t. mIoU by 79.9\% on the S4 subset, 57.9\% on the MS3 subset and 31.2\% on the AVSS subset, respectively. Figure~\ref{fig:intro} illustrates that the inference speed of AVESFormer exceeds previous methods with the ResNet-50 backbone by large margins. In summary, these results demonstrate the advantages of AVESFormer in terms of performance, speed, and model size.

\subsection{Ablation Study}
\paragraph{Training Setup.} We provide ablation results with AVESFormer. To make quick evaluations, we adopt ResNet-50 as the backbone and perform extensive experiments on the S4 and MS3 sub-tasks. Other training settings remain consistent with Section~\ref{sec:implementation}.

\paragraph{Prompt Query Generator}
To verify the effectiveness of our prompt query generator, we remove it to fuse modality with only one audio feature. Additionally, the query generator (QG) in \cite{gao2024avsegformer} and a bias query generator (BQG) are also included. The ordinary query generator follows default settings with 6 layers and 300 queries. The bias query generator replicates the audio query and adds a learnable bias term to it. As shown in Table~\ref{tab:ablation_query_generator}, PQG treats the audio feature as a prompt and cleverly addresses dissipation to avoid attention dissipation, yielding more improvements than the bias query generator. 

\begin{figure}[t]
    \centering
    \includegraphics[width=1.0\textwidth]{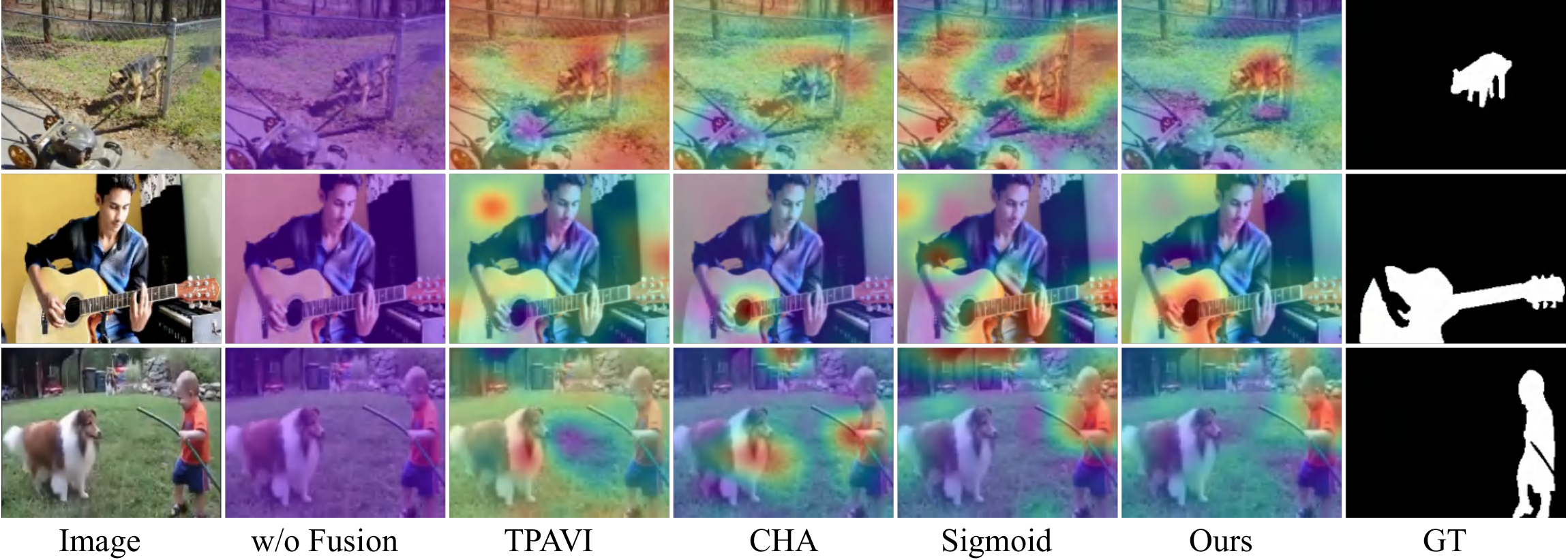}
    \caption{Visualization of attention maps, including no fusion, TPAVI~\cite{zhou2022audio}, channel attention mixer (CHA)~\cite{gao2024avsegformer}, sigmoid attention~\cite{zhou2022acloserlook} and our ELF decoder. Each map shows the correlation between audio queries and visual patches. Red indicates higher attention score while blue indicates lower.}
    \label{fig:attention_map}
\end{figure}

% check the result
\begin{table}[t]
  \footnotesize
  \begin{minipage}{0.48\textwidth}
  \centering
  \setlength{\tabcolsep}{2mm}
  \caption{Effect of the prompt query generator. Prompt query generator overcomes attention dissipation to gain more improvements.
    }
    \begin{tabular}{c cc c cc}
    \toprule
    \multirow{2}{*}{Method} &  \multicolumn{2}{c}{S4} & & \multicolumn{2}{c}{MS3}\\
    \cmidrule{2-3} \cmidrule{5-6}
     & $\mathcal{J}$ & $\mathcal{F}$ & & $\mathcal{J}$ & $\mathcal{F}$\\
     \midrule
    w/o QG&75.9 &87.1 &  & 50.0 & 61.9\\
    QG~\cite{gao2024avsegformer}&78.5 & 88.7 &  & 50.0 & 61.7\\
    BQG & 75.9 & 87.1 && 49.6 & 60.0\\
    \textbf{PQG} & \textbf{79.9} & \textbf{89.1} &  & \textbf{57.9} & \textbf{68.7}\\
    \bottomrule
    \end{tabular}
    \label{tab:ablation_query_generator}
  \end{minipage}
  \hfill
\begin{minipage}{0.48\textwidth}
  \centering
  \setlength{\tabcolsep}{1.6mm}
\caption{Performance of AVSegFormer~\cite{gao2024avsegformer} with and without PQG. S4 may show slight improvement while MS3 shows great improvement after addressing attention dissipation by PQG.
    }
    \begin{tabular}{l cc c cc}
    \toprule
    \multirow{2}{*}{AVSegFormer} &  \multicolumn{2}{c}{S4} & & \multicolumn{2}{c}{MS3}\\
    \cmidrule{2-3} \cmidrule{5-6}
     & $\mathcal{J}$ & $\mathcal{F}$ & & $\mathcal{J}$ & $\mathcal{F}$\\
     \midrule
     w/o PQG & 76.5 & 85.9 && 49.5 & 62.8 \\ 
     \textbf{w/ PQG} & \textbf{77.4} & \textbf{86.9} && \textbf{56.0} & \textbf{67.7} \\ 
     \bottomrule
    \end{tabular}
    \label{tab:ablation_avsegformer}
\end{minipage}
\end{table}

\paragraph{Influence with Plug and Play PQG.}
Furthermore, PQG can be integrated into other models such as AVSegFormer~\cite{gao2024avsegformer}, as shown in Table~\ref{tab:ablation_avsegformer}. Since S4 (single source) is less strict to the audio distinguishing capability, PQG merely exhibits a slight improvement. However, on MS3, where the audio distinguishing capability is crucial due to the presence of multiple sound sources within an image, PQG demonstrates substantial improvement (+6.5\% mIoU) when applied to AVSegFormer.

\paragraph{ELF Decoder.}
We analyze the influence of convolution positioned at different stages of the ELF decoder. As shown in Table~\ref{tab:elf_decoder}, "C" denotes convolution and "T" denotes transformer. The "Stage" column indicates the insertion stage of convolution, with three options listed:  early (C-T-T), middle (T-C-T) and deep (T-T-C). Additionally, a pure transformer decoder (T-T-T) is included. As convolution blocks are moved deeper, the mIoU drops by 2.81\% on S4 and 2.73\% on MS3. This decline can be attributed to the fact that early layers primarily generate local responses. In contrast, deeper layers facilitate high-level interactions between audio-visual modalities, which are essential for AVS tasks.

\paragraph{Fusion Strategy.}
Furthermore, the impact of cross-attention after addressing attention dissipation compared to other fusion strategies is investigated. Four representative fusion strategies are adopted: a) no audio-visual modality fusion, which can be caused by attention dissipation, b) TPAVI proposed in AVSBench~\cite{zhou2022audio}, c) channel attention adopted in AVSegFormer~\cite{gao2024avsegformer}, d) sigmoid attention evoked in CAVP~\cite{zhou2022acloserlook}. Results are shown in Table~\ref{tab:ablation_fusion}. After addressing attention dissipation, our ELF decoder with cross-attention fusion emerges as the optimal choice, demonstrating the most distinguishing representation capability. Figure~\ref{fig:attention_map} shows the attention map visualizations of different fusion strategies.

\begin{figure}[t]
    \centering
    \includegraphics[width=1.0\textwidth]{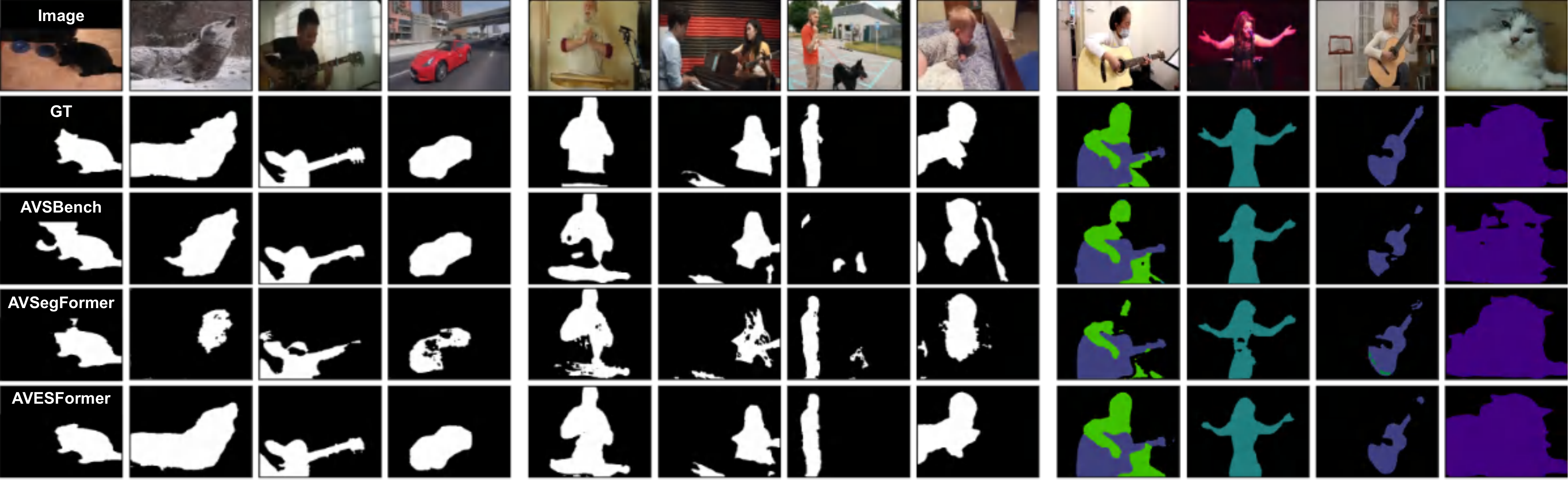}
    \caption{Visualization of segmentation predictions on S4 (left), MS3 (middle) and AVSS (right) Dataset with AVSBench~\cite{zhou2022audio} and AVSegFormer~\cite{gao2024avsegformer}.}
    \label{fig:visual}
\end{figure}

\begin{table}[t]
\footnotesize
\begin{minipage}{0.48\textwidth}
\centering
\setlength{\tabcolsep}{1.6mm}
  \caption{Impact of the convolution blocks at different stages. We show model performance with different convolution insertion stages.}
  \tablestyle{8pt}{1.2}\begin{tabular}{l cc c  cc}
  \toprule
    \multirow{2}{*}{Stage} & \multicolumn{2}{c}{S4} &&  \multicolumn{2}{c}{MS3} \\
    \cmidrule{2-3} \cmidrule{5-6}
     &  $\mathcal{J}$ & $\mathcal{F}$ &&  $\mathcal{J}$ & $\mathcal{F}$ \\
    \midrule
    T-T-T& 77.3 & 87.6 && 56.2 & 66.6 \\
    \textbf{C-T-T}& \textbf{79.9} & \textbf{89.1} && \textbf{57.9} & \textbf{68.7} \\
    T-C-T& 77.6 & 88.0 && 56.5 & 67.3 \\
    T-T-C& 77.1 & 88.3 && 55.2 & 67.3 \\
    \bottomrule
\end{tabular}
  \label{tab:elf_decoder}
  \end{minipage}
  \hfill
  \begin{minipage}{0.48\textwidth}
  \centering
  \setlength{\tabcolsep}{1.6mm}
\caption{Performance of different fusion strategies. It is shown that after fixing attention dissipation, plain cross-attention fusion works better.
    }
    \begin{tabular}{l cc c cc}
    \toprule
    \multirow{2}{*}{Method} &  \multicolumn{2}{c}{S4} & & \multicolumn{2}{c}{MS3}\\
    \cmidrule{2-3} \cmidrule{5-6}
     & $\mathcal{J}$ & $\mathcal{F}$ & & $\mathcal{J}$ & $\mathcal{F}$\\
     \midrule
     w/o fusion & 79.2 & 88.1 && 47.1 & 60.9 \\ 
     w/ TPAVI~\cite{zhou2022audio} & 79.6 & 88.7 && 55.4 & 65.4 \\ 
     w/ CHA~\cite{gao2024avsegformer} & 79.6 & 88.6 && 55.7 & 65.8 \\ 
     w/ sigmoid~\cite{zhou2022acloserlook} & 78.4 & 88.6 && 55.3 & 62.0\\
     \textbf{w/ ELF} & \textbf{79.9} & \textbf{89.1} && \textbf{57.9} & \textbf{68.7} \\ 
        \bottomrule
    \end{tabular}
    \label{tab:ablation_fusion}
\end{minipage}
\end{table}

\begin{wraptable}{rh}{0.5\textwidth}
    \centering
    \caption{Performance of the number of queries. 
    }
    
    \tablestyle{8pt}{1.05}\begin{tabular}{c|cc|cc}
        % \toprule
    	\multirow{2}{*}{} &  \multicolumn{2}{c|}{S4} &  \multicolumn{2}{c}{MS3}\\

     \# of queries& $\mathcal{J}$ & $\mathcal{F}$ & $\mathcal{J}$ & $\mathcal{F}$\\
     \midrule[1pt]
    	8 & 79.3 & 88.9 & 55.8 & 66.0 \\ 
    	\textbf{16} & \textbf{79.9} & \textbf{89.1} & \textbf{57.9} & \textbf{68.7} \\ 
    	32 & 79.4 & 88.9 & 56.2 & 66.6 \\ 
    	64 & 79.1 & 88.9 & 55.8 & 67.0 \\ 
     128 & 79.0 & 88.8 & 56.0 & 67.4 \\
     256 & 79.3 & 89.0 & 57.3 & 67.8 \\
        % \bottomrule
    \end{tabular}
    \label{tab:ablation_nq}
\end{wraptable}

\paragraph{Number of Queries.}
Table~\ref{tab:ablation_nq} presents the results of AVESFormer trained with varying numbers of quires of AVS dataset. Experiments are conducted with query numbers ranging from 8 to 256 with a scale factor of 2. Notably, using 16 queries performs best across S4 and MS3. This suggests that even though there are a number of sounding object categories, a large number of queries may not be necessary. A few queries in AVESFormer are adequate for learning distinguishing audio features.

\paragraph{Qualitative Analysis.}
Visualizations of AVESFormer compared with those of AVSBench~\cite{zhou2022audio} and AVSegFormer~\cite{gao2024avsegformer} in ResNet-50 backbone on AVSBench-object and AVSBench-semantic datasets are depicted in Figure~\ref{fig:visual}. Our AVESFormer overcomes critical attention dissipation and makes more sophisticated visualization and segmentation performance. See Appendix~\ref{app:qualitative_analysis.} for more visualizations.

\section{Conclusion and Discussion}
\paragraph{Conclusion.} In this paper, we analyze the attention dissipation phenomenon and inefficient transformer decoder. Based on these findings, we introduce AVESFormer, the first transformer-based real-time AVS model. Experimental results demonstrate that AVESFormer achieves the new state-of-the-art performance-speed trade-off. We hope our method provides insights into new architecture design not only in AVS tasks but also in various multi-modality scenarios. 

\paragraph{Limitation and Future Work.} There still exist limitations on AVESFormer. On one hand, the audio backbone Vggish~\cite{hershey2017vggish} constitutes about 60\% of the model parameters, posing challenges for deployment on mobile devices. On the other hand, temporal information is ignored in real-time AVS scenario. These will be the focus of our future work.
\newpage

\bibliography{sample}
%%%%%%%%%%%%%%%%%%%%%%%%%%%%%%%%%%%%%%%%%%%%%%%%%%%%%%%%%%%%

\clearpage
\appendix

\section*{Appendix}

\section{Attention Dissipation}
\subsection{Proof on Attention Dissipation}\label{app:dissipation_detail}
As discussed in Sec. \ref{sec:dissipation}, a brief explanation of attention dissipation is given. Now, we will provide more detailed proof of this phenomenon.

As commonly practised in AVS tasks, visual features are extracted from the visual backbone to get $\mathcal{F}_{visual}\in\mathbb{R}^{c\times h\times w}$ of one frame. Then we patchify the visual feature into $\mathcal{P}_{visual}\in\mathbb{R}^{N\times c}$ where $N=h\times w$. Meanwhile, audio signals within one frame are input into the audio backbone to form $\mathcal{F}_{audio}\in\mathbb{R}^{1\times c}$. Note that since we only consider one frame at a time in real-time scenario, the sequence length of the audio feature is equal to 1. We cannot omit the sequence length dimension because we should keep this shape to perform matrix multiplication in the attention mechanism.

Consequently, the modality fusion process is performed originally by cross attention, where visual patches are query while the audio feature is key and value:
\begin{align}
    O&=\text{Softmax}(\mathcal{P}_{visual}\mathcal{F}_{audio}^T)\mathcal{F}_{audio}\in\mathbb{R}^{N\times c},
\end{align}
where
\begin{align}
    \mathcal{P}_{visual}&=\begin{bmatrix}
        q_1\\
        q_2\\
        \vdots\\
        q_N
    \end{bmatrix},\\[5pt]
    q_i&\in\mathbb{R}^{1\times c},\quad i\in[1,2,\dots,N],\\[5pt]
    \mathcal{F}_{audio}&=k=v\in\mathbb{R}^{1\times c}.
\end{align}
The attention logit matrix $\mathcal{A}$ can be written as:
\begin{align}
    \mathcal{A}=\mathcal{P}_{visual}\mathcal{F}_{audio}^T
    =&\begin{bmatrix}
        q_1\\
        q_2\\
        \vdots\\
        q_N
    \end{bmatrix}k^T
    =\begin{bmatrix}
        q_1k^T\\
        q_2k^T\\
        \vdots\\
        q_Nk^T
    \end{bmatrix}\in\mathbb{R}^{N\times 1},
\end{align}
where
\begin{align}
    q_ik^T&\in\mathbb{R},\quad i\in[1,2,\dots,N].
\end{align}
Softmax is calculated along the row vector on attention matrix $\mathcal{A}$ to get attention probability matrix $\mathcal{P}$:
\begin{align}
    \mathcal{P}=\text{Softmax}(\mathcal{A})|_{\text{row}}
    =\begin{bmatrix}
        e^{q_1k^T}/\sum e^{q_1k^T}\\
        e^{q_2k^T}/\sum e^{q_2k^T}\\
        \vdots\\
        e^{q_Nk^T}/\sum e^{q_Nk^T}
    \end{bmatrix}
    &=\begin{bmatrix}
        e^{q_1k^T}/e^{q_1k^T}\\
        e^{q_2k^T}/e^{q_2k^T}\\
        \vdots\\
        e^{q_Nk^T}/e^{q_Nk^T}
    \end{bmatrix}
    =\begin{bmatrix}
        1\\
        1\\
        \vdots\\
        1
    \end{bmatrix}
    =\mathbf{1}_{N\times 1}.
\end{align}
Finally the output $\mathcal{O}$ becomes a simply replication of value matrix:
\begin{align}
    \mathcal{O}=\text{Softmax}(\mathcal{A})|_{\text{row}}\mathcal{F}_{audio}
    =\mathcal{P}\mathcal{F}_{audio}
    =\mathbf{1}_{N\times 1}\mathcal{F}_{audio}
    =\begin{bmatrix}
        1\\
        1\\
        \vdots\\
        1
    \end{bmatrix}\mathcal{F}_{audio}
    =\begin{bmatrix}
        \mathcal{F}_{audio}\\
        \mathcal{F}_{audio}\\
        \vdots\\
        \mathcal{F}_{audio}
    \end{bmatrix}.
\end{align}
The attention dissipation phenomenon shows that cross-attention with visual features such as query and audio as key and value turns out to be a simple replication of audio signals. It goes against our original intent of modality fusion.

\subsection{Code implementation}
To make a fully comprehensive understanding of attention dissipation, we provide a PyTorch-like pseudo-code for easy verification and implementation of cross-attention dissipation. Algorithm~\ref{alg:code} provides the pseudo-code of attention dissipation in the AVS task. For the current frame, we calculate the attention matrix with the use of visual features as query and audio as key and value. 

\begin{algorithm}[htbp]
\caption{Pseudo-code of Attention Dissipation in a PyTorch-like style.}
\label{alg:code}

\definecolor{codeblue}{rgb}{0.25,0.5,0.5}
\definecolor{codegray}{rgb}{0.5,0.5,0.5}
\lstset{
  backgroundcolor=\color{white},
  basicstyle=\fontsize{7.2pt}{7.2pt}\ttfamily\selectfont,
  columns=fullflexible,
  breaklines=true,
  captionpos=b,
  commentstyle=\fontsize{7.2pt}{7.2pt}\color{codeblue},
  keywordstyle=\fontsize{7.2pt}{7.2pt}\color{magenta},
  numberstyle=\tiny\color{codegray},
%  frame=tb,
}
\begin{lstlisting}[language=python]
# image, audio: visual and audio feature
# attn: attention matrix
# out: output of attention

import torch
import torch.nn as nn
import torch.nn.functional as F

def cross_attention(image:torch.Tensor, audio:torch.Tensor):
    """
    :param image: torch.tensor with shape [B, C, H, W]
    :param audio: torch.tensor with shape [B, C]
    :return: fused feature and attention weight 
    """
    
    image = image.flatten(2).transpose(1, 2)
    audio = audio.unsqueeze(1)
    
    q = image
    k = audio
    v = audio
    
    attn = torch.matmul(q, k.transpose(1, 2))
    attn = F.softmax(attn, dim=-1)
    out = torch.matmul(attn, v)
    
    return out,attn
\end{lstlisting}
\end{algorithm}
\section{Experiments}
\subsection{Experimental Details}\label{app:exp_detail}
During training, we use the original image size as 224$\times$224. We apply horizontal flipping on S4 and MS3 for data augmentation. Since the S4 sub-set only contains annotations on the first frame in the training split, we only use the first frame to provide supervision. We use the AdamW optimizer and a polynomial learning rate decay with power = 0.9. On S4 and MS3, the learning rate is set to 0.0005, and on AVSS, it is set to 0.0001. Following previous practice \cite{gao2024avsegformer}, we train MS3 for 60 epochs since it is relatively small, while the S4 and AVSS subsets are trained for 30 epochs. Batch size is set to 16 for S4 and MS3 and 8 for AVSS. We adopt two ResNet \cite{he2016resnet} backbones (ResNet-50 and ResNet-18) for the segmentation network. For the audio backbones, we use VGGish \cite{hershey2017vggish} frozen during the training. The prompt query generator (PQG) receives the feature from the audio backbone as prompt. The number of queries is set to 16, and the number of layers is set to 3. At the output end, the audio feature prompt is discarded. The transformer decoder is adopted from Multi-Scale Deformable (MSDeform) attention \cite{zhu2020deformabledetr}. The first two attention blocks are replaced by convolution to form ELF decoder. Convolution blocks are attached with residual connection and LayerNorm \cite{yu2024metaformer}. As for the segmentation loss, on S4 and MS3, we set $\lambda_\text{IoU}=1.8$ and on AVSS  $\lambda_\text{IoU}=1.0$ with $\lambda_\text{Dice}=1.0$ and $\lambda_\text{aux}=0.1$. For inference, since the end-to-end real-time scenario does not support inferring on a bunch of frames (because we want to segment one image at a time on the device), the latency of all models is measured under one single frame, that is, $T=1$. Nevertheless, some of the methods employ temporal information within multiple frames, which would be lost in a single frame scenario; we still keep their performance the same for comparison.

\section{Qualitative analysis}
\subsection{Results Visualization}\label{app:qualitative_analysis.}
\begin{figure}[htbp]
    \begin{minipage}[t]{1.0\linewidth}
        \centering
        \includegraphics[width=1.0\textwidth]{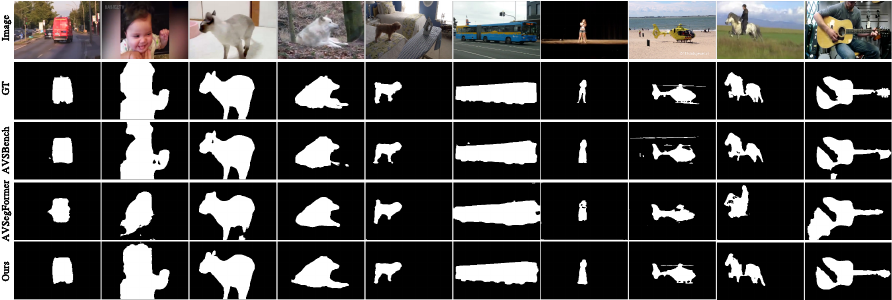}
    \end{minipage}
    \caption{Qualitative audio-visual segmentation results on AVSBench-Object S4 sub-set~\cite{zhou2023avss} by TPAVI~\cite{zhou2022audio}, AVSegFormer~\cite{gao2024avsegformer}, and AVESFormer. Each row represents the raw image, ground truth or different methods. Each column represents various data samples.}
    \label{fig:app_visualization_s4}
    \vspace{0.7cm}
    \begin{minipage}[t]{1.0\linewidth}
        \centering
        \includegraphics[width=1.0\textwidth]{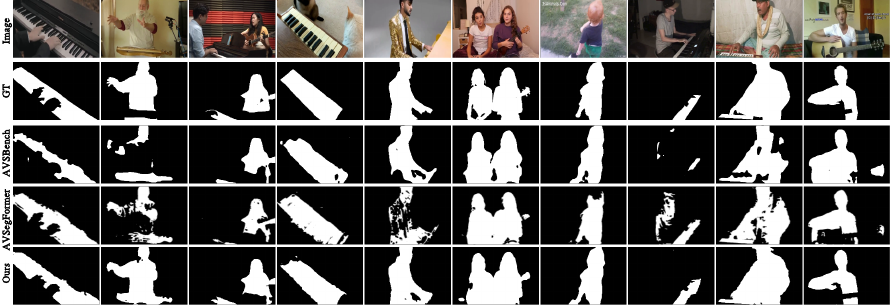}
    \end{minipage}
    \caption{Qualitative audio-visual segmentation results on AVSBench-Object MS3 sub-set~\cite{zhou2023avss} by TPAVI~\cite{zhou2022audio}, AVSegFormer~\cite{gao2024avsegformer}, and AVESFormer. Each row represents the raw image, ground truth or different methods. Each column represents various data samples.}
    \label{fig:app_visualization_ms3}
    \vspace{0.7cm}
    \begin{minipage}[t]{1.0\linewidth}
        \centering
        \includegraphics[width=1.0\textwidth]{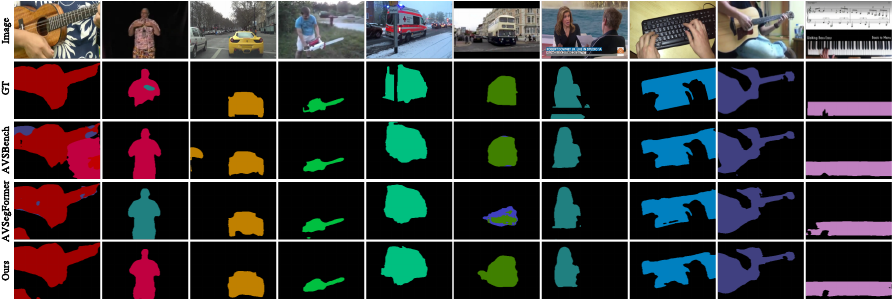}
    \end{minipage}
    \caption{Qualitative audio-visual segmentation results on AVSBench-Semantics~\cite{zhou2023avss} by TPAVI~\cite{zhou2022audio}, AVSegFormer~\cite{gao2024avsegformer}, and AVESFormer. Each row represents the raw image, ground truth or different methods. Each column represents various data samples.}
    \label{fig:app_visualization}
    % \vspace{-2em}
\end{figure}

We present additional visualization results for the paper, alongside AVSBench~\cite{zhou2022audio}, AVSegFormer~\cite{gao2024avsegformer} and our model on AVSBench-Object~\cite{zhou2022audio} and AVSBench-Semantic~\cite{zhou2023avss} with ResNet-50~\cite{he2016resnet} backbone, as depicted in Figure.~\ref{fig:app_visualization_s4}, Figure.~\ref{fig:app_visualization_ms3}, and Figure.~\ref{fig:app_visualization}. We demonstrate that AVESFormer efficiently presents a more fine-grained prediction and a more accurate audio-visual corresponding capability to the segmentation of objects in the scene compared to previous methods.

\subsection{Attention Map in Decoder Blocks}\label{app:qualitative_early.}
\begin{figure}[htbp]
\vspace{-0.2cm}
    \centering
    \includegraphics[width=0.75\linewidth]{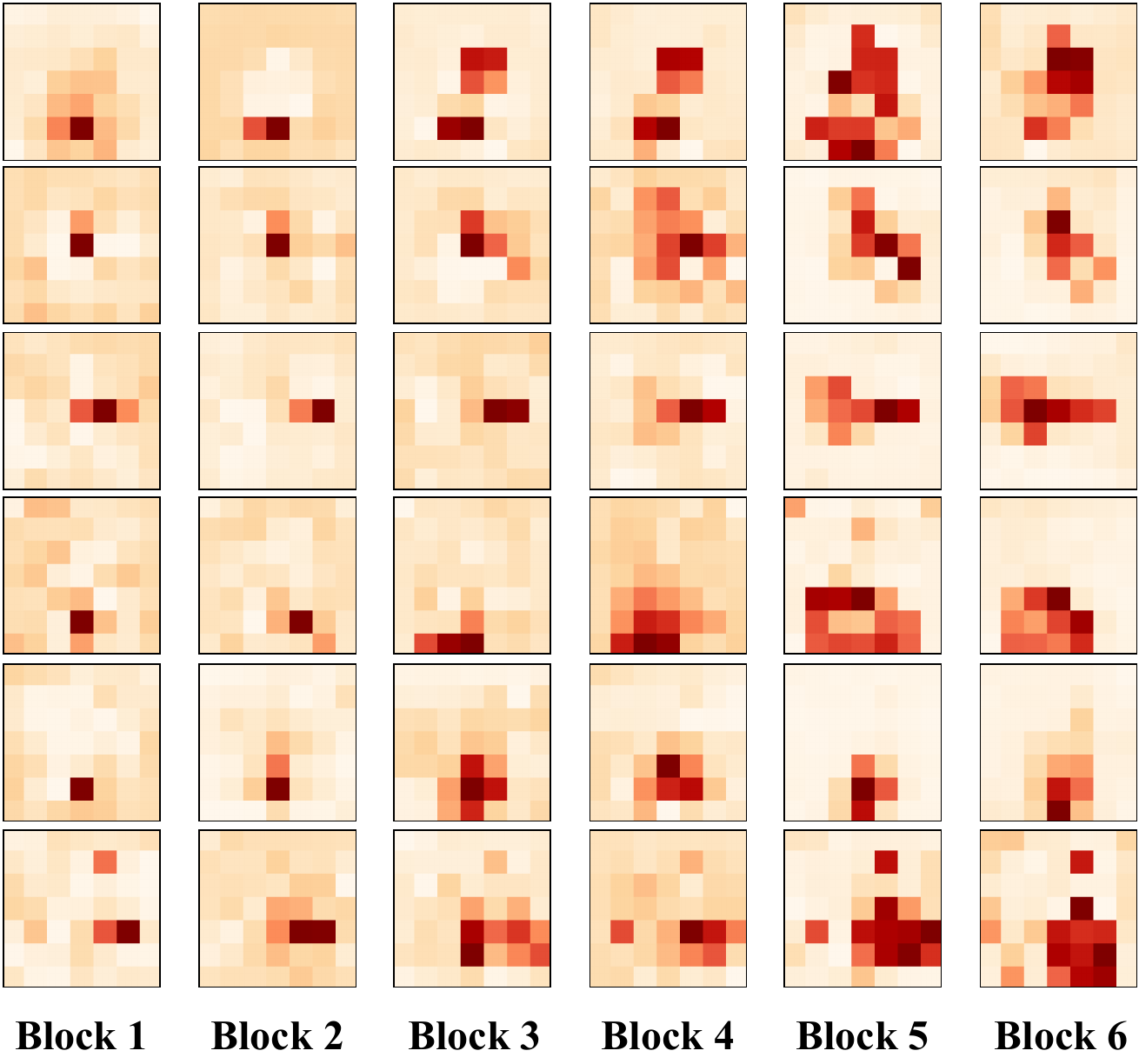}
    \caption{Attention probabilities of a full transformer decoder. Each map shows the attention probability of a query audio feature to all visual patches. Maps are averaged along all heads. Each row represents an image sample in the test set. Each column represents a decoder block. Dark red indicates higher attention probability and shallow orange indicates low attention probability.}
    \label{fig:early_attention}
    % \vspace{-2em}
\end{figure}

We present the visualization of a full transformer decoder architecture to show the attention pattern of different stages. As shown in Figure. \ref{fig:early_attention}, the illustration represents a narrow local attention map in the early stages. As the stage goes deeper, the attention region grows larger gradually. In the end, the attention region becomes large and fine-grained, which is suitable for the segmentation task.

\section{Broader Impacts}
Our AVESFormer has taken a step towards real-time performance in AVS tasks. Our model has seen significant advancements, allowing for applications across various domains. Potentially, our model may bring positive impact on reducing computational cost, improving quality of life and enhancing automation and efficiency. Negative impact may lies on surveillance, privacy concerns and the dependence on data quality. Addressing these challenges is crucial for the responsible and ethical deployment of these models.

\end{document}